# SpineReport: Automated 3D Quantification and Reporting of Lumbar Spine Degeneration on MRI

Nathan Molinier[1,2], Adrian A. Marth[3,4], Reto Sutter[3,4], Christoph Germann[3,4], Jacob A. Connolly[5,6], Mathieu Guay-Paquet[1], Nathan D. Schilaty[5,6,7], Kenneth A. Weber II[8], Julien Cohen-Adad[1,2,9,10]

[1]*NeuroPoly, Institute of Biomedical Engineering, Polytechnique Montreal, Montreal, QC, Canada*
[2]*Mila, Quebec AI Institute, Montreal, QC, Canada*
[3]*Department of Radiology, Balgrist University Hospital, Zurich, Switzerland*
[4]*Faculty of Medicine, University of Zurich, Zurich, Switzerland*
[5]*Department of Neurosurgery, Brain & Spine, University of South Florida, Tampa, FL, USA*
[6]*Center for Neuromusculoskeletal Research, University of South Florida, Tampa, FL, USA*
[7]*Department of Medical Engineering, University of South Florida, Tampa, FL, USA*
[8]*Stanford School of Medicine, Stanford, CA, USA*
[9]*Functional Neuroimaging Unit, CRIUGM, Université de Montréal, Montreal, QC, Canada*
[10]*Centre de recherche du CHU Sainte-Justine, Université de Montréal, Montreal, QC, Canada*

Corresponding Author: Julien Cohen-Adad, Phd; MSc; Eng
Email: julien.cohen-adad@polymtl.ca

# Abstract

Lumbar spine conditions are a leading cause of disability worldwide, yet reliable quantification of degeneration from MRI remains challenging. In clinical practice, analysis is predominantly performed in two dimensions (2D), as manual three-dimensional (3D) assessment is time-consuming. However, 2D measurements suffer from limited reproducibility, particularly when anatomical structures are not aligned with the imaging plane. Existing automated approaches are often restricted to 2D, rely on discrete grading, or lack robustness and interpretability.

We introduce *SpineReport*, an open-source, fully automated framework for comprehensive 3D morphometric analysis of lumbar spine MRI. Leveraging robust anatomical segmentations, the method extracts quantitative metrics from key structures, including the spinal canal, spinal cord, vertebrae, intervertebral discs, and foramina. These include both morphological and signal-based features, enabling cross-subject and longitudinal assessment. *SpineReport* further generates subject-specific reports that allow comparison with cohort distributions, improving interpretability and objective characterization of spinal morphology.

Clinical relevance was evaluated against radiologist-reported severity grades for central canal, lateral recess, and foraminal stenosis. Metrics showed strong associations with central canal stenosis severity, with T2-weighted CSF signal providing the highest performance (AUC = 0.95). Canal AP diameter and area ratios also demonstrated strong correlations and high discriminative ability (AUC > 0.80). For lateral recess stenosis, associations were moderate, with lateral CSF signal being the most informative (AUC = 0.73). No significant associations were observed for foraminal stenosis despite robust region-of-interest extraction.

*SpineReport* is released as a standalone, open-access tool, facilitating integration into research and clinical workflows: https://ivadomed.github.io/SpineReport/

# 1. Introduction

Lumbar spine conditions are the leading causes of disability worldwide, representing a major public health burden across all age groups (Hartvigsen et al. 2018). Low back pain (LBP) is a broad clinical term encompassing pain, muscle tension, or stiffness localized below the costal margin and above the inferior gluteal folds, with or without associated leg pain such as sciatica. It can arise from a wide range of pathological conditions affecting the lumbar spine such as degenerative changes, infection, osteoporosis, inflammatory disease, and fractures. However, despite extensive diagnostic workups, the majority of cases are ultimately classified as non-specific low back pain (Maher et al. 2017).

Magnetic resonance imaging (MRI) provides excellent soft-tissue contrast and detailed visualization of spinal anatomy, making it the reference imaging modality for the clinical evaluation of lumbar spine conditions, especially LBP (Samartzis et al. 2013). MRI enables the assessment of disc degeneration, spinal canal stenosis, vertebral abnormalities, edema, tumors and neural compression. Nevertheless, the interpretation of lumbar spine MRI remains challenging, with substantial intra- and inter-rater variability reported in the evaluation of degenerative findings and their clinical relevance (Herzog et al. 2017).

In current clinical practice, quantitative measurements, if taken, are often performed manually on two-dimensional (2D) slices. Such measurements are sensitive to slice orientation and subject positioning, and may be prone to errors when anatomical structures are misaligned with the imaging plane. The three-dimensional morphology of the spinal canal, the spinal cord, vertebrae, intervertebral foramina (IVF) and intervertebral discs (IVD), cannot be reliably captured from a small number of 2D views. Tunset et al. (2013) highlighted that inter-rater variability arises from selecting different sagittal 2D slices for disc measurements, leading to inconsistencies in the results.

To address these limitations, numerous automatic and semi-automatic methods have been proposed for lumbar spine analysis. Existing approaches include shape deformation models for lumbar spine reconstruction based on mid-sagittal slices (Qian et al. 2025), as well as pipelines combining 2D segmentation with classification models to grade disc herniation, canal stenosis, or foraminal stenosis at predefined spinal levels (LewandrowskI et al. 2020). More recent methods extend these ideas by integrating 2D segmentations with measurement extraction and large language models for automated report generation (Salem et al. 2025). Deep learning–based systems have also been developed to predict semi-quantitative grades or measurements of disc degeneration, disc narrowing, central canal stenosis, endplate defects, marrow changes, IVF stenosis, spondylolisthesis, and disc herniation (Hallinan et al. 2021; Zheng et al. 2022; Windsor et al. 2022; Windsor et al. 2024; Wang et al. 2025; Ghobrial and Roth 2025; Subramanian et al. 2025; Purushottam et al. 2026; Udomluck et al. 2026).

Despite their promise, existing tools share several important limitations. Many methods rely exclusively on 2D analysis, restricting their robustness in the presence of anatomical variability, spinal pathology or scan misalignment. Most approaches focus on a limited subset of structures or pathologies and produce discrete severity grades rather than continuous, quantitative measurements of structures of interest. Their performance often degrades on external datasets due to limited robustness across imaging contrasts, resolutions, and fields of view. Furthermore, measurement strategies are frequently insufficiently described, and many methods are not publicly available, limiting reproducibility and clinical translation.

Quantitative extraction of lumbar spine morphometrics can offer objective characterization of spinal degeneration, including spinal canal and IVF stenosis, disc pathology, vertebral body height reduction, and nerve root compression. Changes in spinal canal and IVF shape have been shown to correlate with nerve root compression and radiculopathy symptoms (Bartynski and Petropoulou 2007; Gavotto et al. 2025). Similarly, intervertebral disc height and volume are associated with pain scores and are commonly used to evaluate treatment outcomes, including surgical and non-surgical interventions (X. Chen et al. 2021). Yet, these metrics of spinal canal / IVF shape and intervertebral disc height are rarely reported on neuroradiological reports. Widely used grading systems such as the Pfirrmann and modified Pfirrmann scores are themselves based on disc morphology and signal characteristics, underscoring the importance of quantitative structural assessment (Pfirrmann et al. 2001; Griffith et al. 2007).

In this work, we present *SpineReport*, an open-source and robust framework for the automatic extraction of full-spine 3D morphometrics from MRI. Building upon 3D anatomical segmentation tools such as *TotalSpineSeg* (Warszawer et al. 2025), the proposed pipeline derives quantitative morphometric features from key spinal structures, including the spinal canal, the spinal cord, the cauda equina, vertebrae, IVF, and IVD. These measurements are aggregated into comprehensive, personalized reports that enable rapid comparison between individual patients and reference populations. We demonstrate the clinical utility of *SpineReport* by benchmarking automated measurements of spinal canal, lateral recess, and IVF stenosis against expert radiological severity assessments.

# 2. Materials and Methods

## 2.1 Framework overview

*SpineReport* is designed to extract quantitative morphometric measurements from spinal anatomy segmentations, that include the spinal canal, the spinal cord, vertebrae, IVF and IVD. Importantly, vertebrae and IVD segmentations must be multiclass, such that each anatomical structure (e.g., vertebrae L2 and L3) is assigned a distinct label. Here, we used *TotalSpineSeg* (Warszawer et al. 2025) to generate the segmentation used by *SpineReport*, given its robustness across a wide range of MRI acquisition protocols (contrast, field of view, spatial resolution, missing vertebrae, image artifacts), but other tools (van der Graaf et al. 2024; J. Chen et al. 2024; Möller et al. 2025) could in principle be used instead.

Prior to morphometric analysis, all MRI scans and corresponding segmentations are resampled to a common isotropic resolution of 1 $mm^3$ and reoriented to the LAS+ coordinate convention. Working in a 1 $mm^3$ isotropic space provides two key advantages: (i) *TotalSpineSeg* produces more detailed, upsampled anatomical segmentations, and (ii) isotropic voxel spacing simplifies the extraction and interpretation of morphometric measurements expressed in physical units (e.g., millimeters and square millimeters).

## 2.2 Processing pipeline

The proposed processing pipeline consists of two main stages, as illustrated in **Figure 1**. First, *TotalSpineSeg* is applied to the 3D MRI scans to automatically segment the spinal anatomy and identify individual instances of the spinal canal, the spinal cord, vertebrae and IVD.

Second, quantitative morphometric metrics are extracted by iterating over all segmented anatomical structures. This step derives morphological-based measurements and, when applicable, signal-based features specific to cerebrospinal fluid (CSF) and IVD, leveraging water-sensitive MRI sequences such as T2-weighted imaging. A more detailed description of the extracted metrics is provided in the following section.

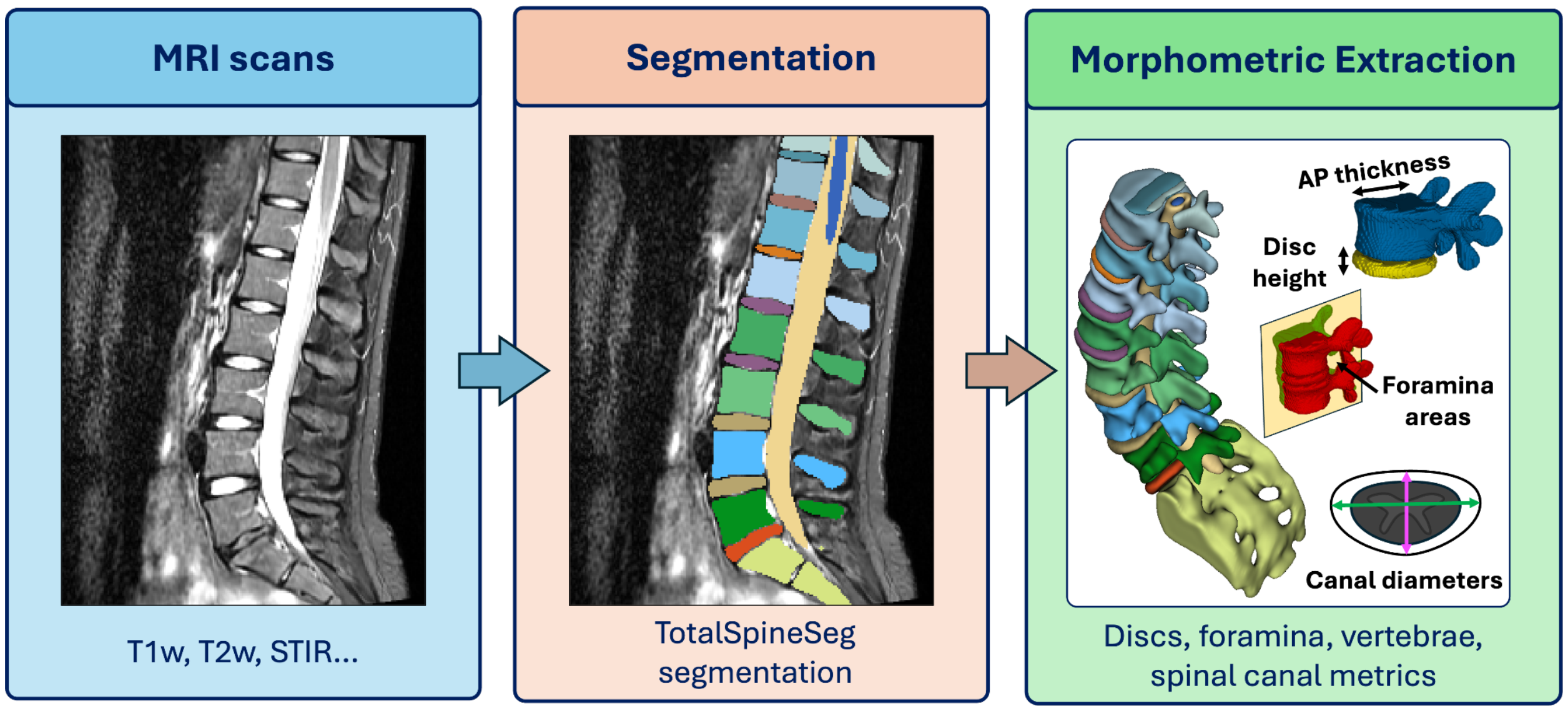

**Figure 1**. Overview of the SpineReport morphometric analysis pipeline. The preprocessed MRI scan is segmented using TotalSpineSeg, after which morphometric measurements are automatically and iteratively extracted from all segmented spinal structures, including the spinal canal, the spinal cord, vertebrae, IVF, and IVD.

## 2.3 Morphometrics extraction

This section describes the anatomical reference frames used for morphometric extraction and provides detailed figures and explanations illustrating how morphometric measurements are derived from the 3D segmentations.

### 2.3.1 Anatomical references

To facilitate morphometric extraction, to ensure inter-subject and between-sequence consistency, and to maximize repeatability across multiple acquisitions of the same subject, several global anatomical reference frames are defined for the different spinal structures.

First, following the approach implemented in the *Spinal Cord Toolbox* (De Leener et al. 2017), a spinal canal centerline is derived from the initial canal segmentation by calculating the center of mass for each axial slice. To approximate the canal's local orientation and curvature, a smooth centerline is generated via B-spline interpolation. This centerline is defined in world coordinates, with tangent vectors extracted at each sampling point to represent the spinal canal's trajectory.

In addition, a spine centerline is extracted to better represent the alignment of the vertebral column. This centerline is obtained by aggregating the vertebral bodies and IVD into a single binary segmentation, computing the slice-wise center of mass, and applying interpolation to produce a smooth trajectory centered on the vertebral bodies similarly to the canal centerline. In addition to the canal centerline, this reference defines the superior–inferior (SI), anterior–posterior (AP), and right–left (RL) axes for all structures, ensuring consistent measurements across subjects—even in cases of pronounced spinal curvature (e.g., scoliosis)—by establishing a straightened coordinate system.

Finally, point-wise intervertebral labels located at the posterior margin of each IVD are used as anatomical landmarks. These landmarks enable reliable tracking of subject size and spinal extent, and are particularly useful for normalizing and scaling measurements across patients during report generation.

### 2.3.2 Spinal canal and spinal cord

Spinal canal morphometrics, shown in **Figure 2**, are computed slice-wise along the superior–inferior axis using 2D cross-sections orthogonal to the spinal canal centerline. At each index, the tangent vector of the centerline defines an orthogonal plane from which cross-sectional area, AP and RL diameters, eccentricity, and solidity are derived. The same measurements are computed for the spinal cord using its corresponding segmentation.

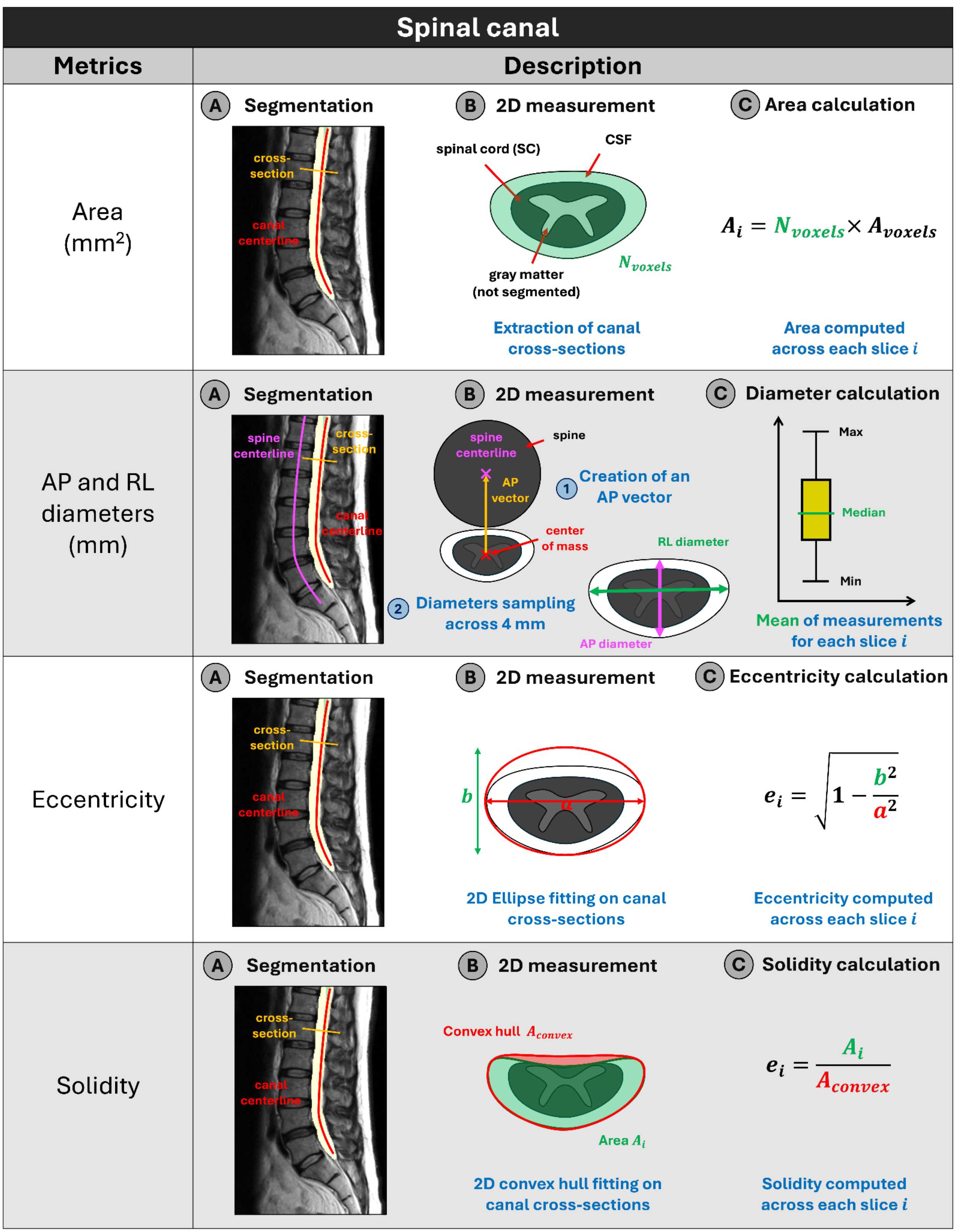

**Figure 2.** Morphological measurements of the spinal canal illustrating how the cross-sectional area, AP and RL diameters, eccentricity, and solidity are derived from the segmentation.

In addition to morphological measurements, CSF signal was extracted on T2-weighted MRI using the same cross-sections as for canal analysis. As *TotalSpineSeg* does not explicitly segment CSF, the canal-minus-cord region also includes neural elements (e.g., cauda equina). To approximate CSF signal, the 90th percentile intensity within segmented pixels was extracted per slice and normalized by the 90th percentile across the full segmentation to enable inter-subject comparison. Additionally, the 10% most lateral pixels on each side were used to derive lateral CSF signal (left and right) estimates per slice. While these signal features provide valuable information, they remain relative measures, as T2-weighted images are qualitative and influenced by factors such as field inhomogeneity, sequence variability, and partial volume effects. To avoid misinterpretation, comparisons in this study are restricted to scans acquired with similar sequences and assuming similar scanning conditions.

### 2.3.3 Vertebrae

Vertebrae are analyzed individually, with measurements restricted to the vertebral body. Specifically, AP thickness, SI thickness, and vertebral body volume are extracted as shown in **Figure 3**. To isolate the vertebral body from the vertebral processes, a cutting plane is defined using the spinal canal centerline, along with a symmetry axis obtained by projecting the vertebra onto a plane orthogonal to the centerline tangent.

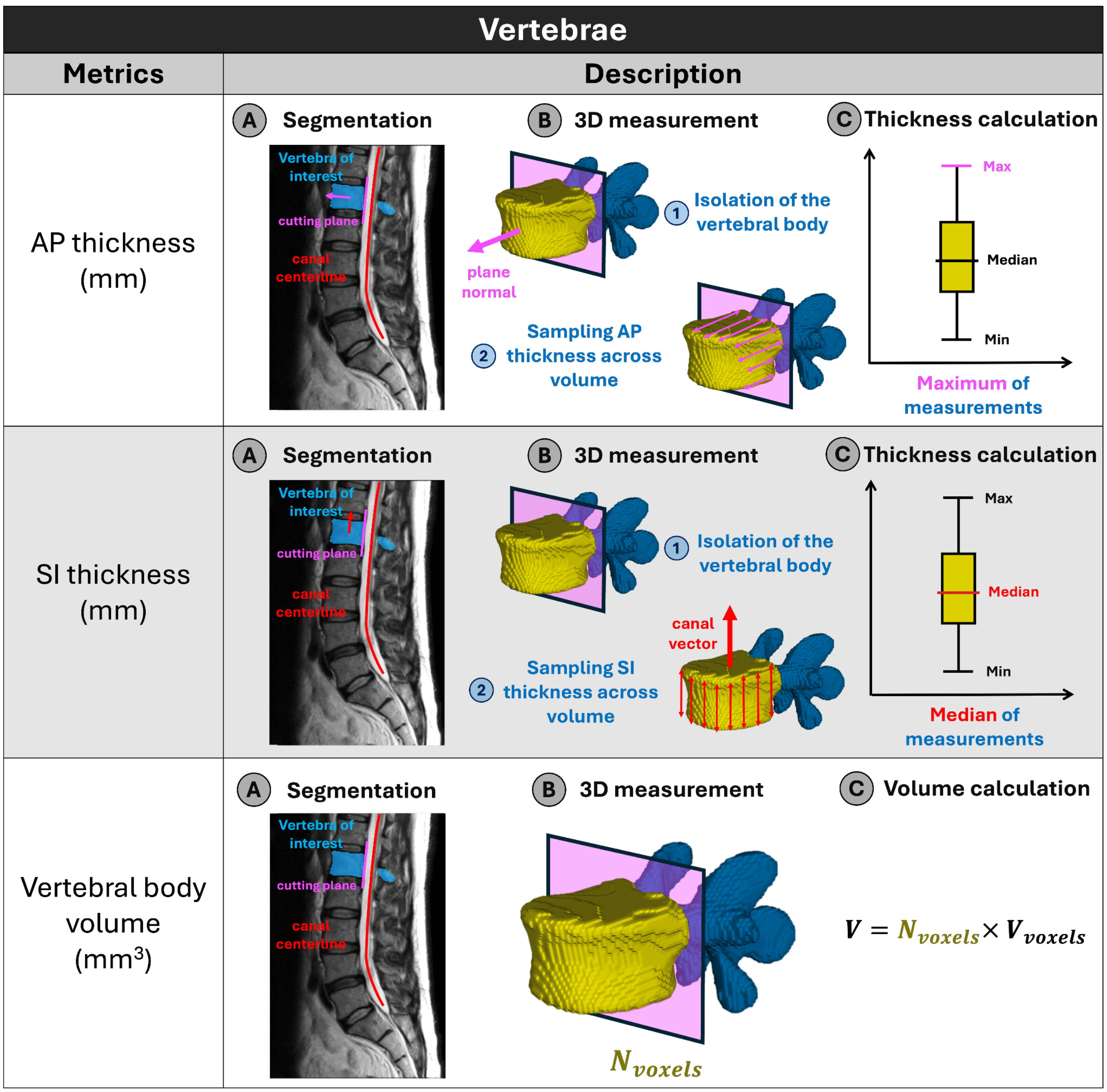

**Figure 3.** Morphological measurements of the vertebrae illustrating how the volume, the AP and the RL thickness are derived from the segmentation.

## 2.3.4 Intervertebral foramina

The IVF is the opening formed by two adjacent vertebrae and the intervening disc, serving as a key pathway for neurovascular structures between the spinal canal and peripheral compartments. Unlike other foramina (e.g., the foramen magnum), it is not a fully enclosed 3D bony structure due to the mobility of the ventral intervertebral and dorsal zygapophysial joints; its size can vary with pathology and subject positioning, and it is present bilaterally (left and right) (Gilchrist et al. 2002).

To extract the foramina at a given level, the segmentations of the superior and inferior vertebrae and the IVD are combined. A symmetry plane is defined using the spinal canal centerline tangent at the disc level, and an AP axis is derived from the canal and spine centerline points. This enables separation of the right and left foramina. Since the foramina are not fully enclosed, each segmentation half is projected onto the symmetry plane to estimate its lateral area and compute a RL asymmetry ratio as shown in **Figure 4**.

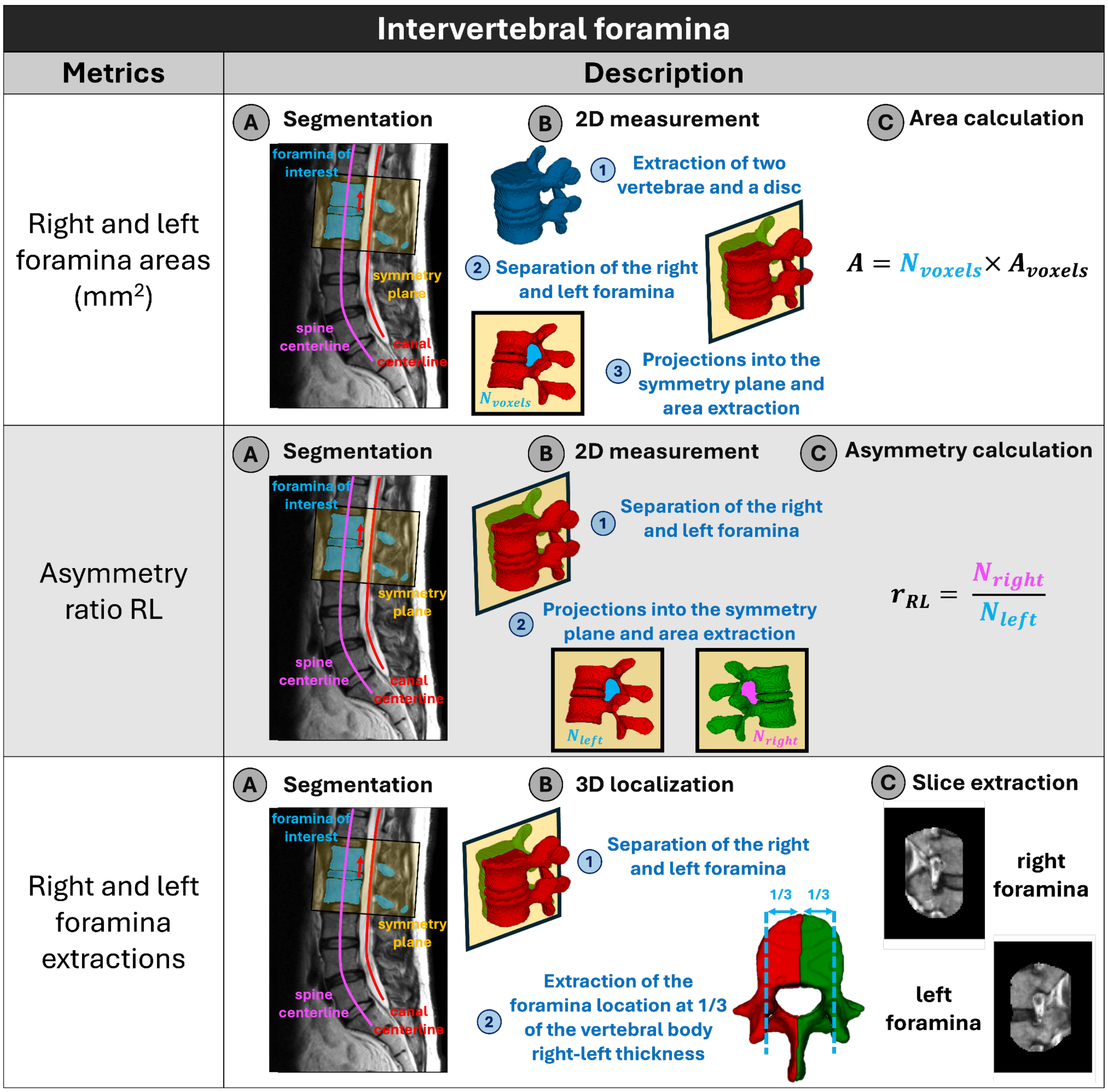

**Figure 4.** Morphometric extraction of the IVF illustrating how left and right areas, asymmetry, and the foraminal slice are derived from the segmentation and image.

### 2.3.5 Intervertebral discs

Intervertebral discs (IVD) are analyzed individually, with morphological measurements extracted including disc height (DH), disc height index (DHI), volume, eccentricity, and solidity as shown in **Figure 5**. The SI axis is defined by the tangent vector of the centerline at the disc level.

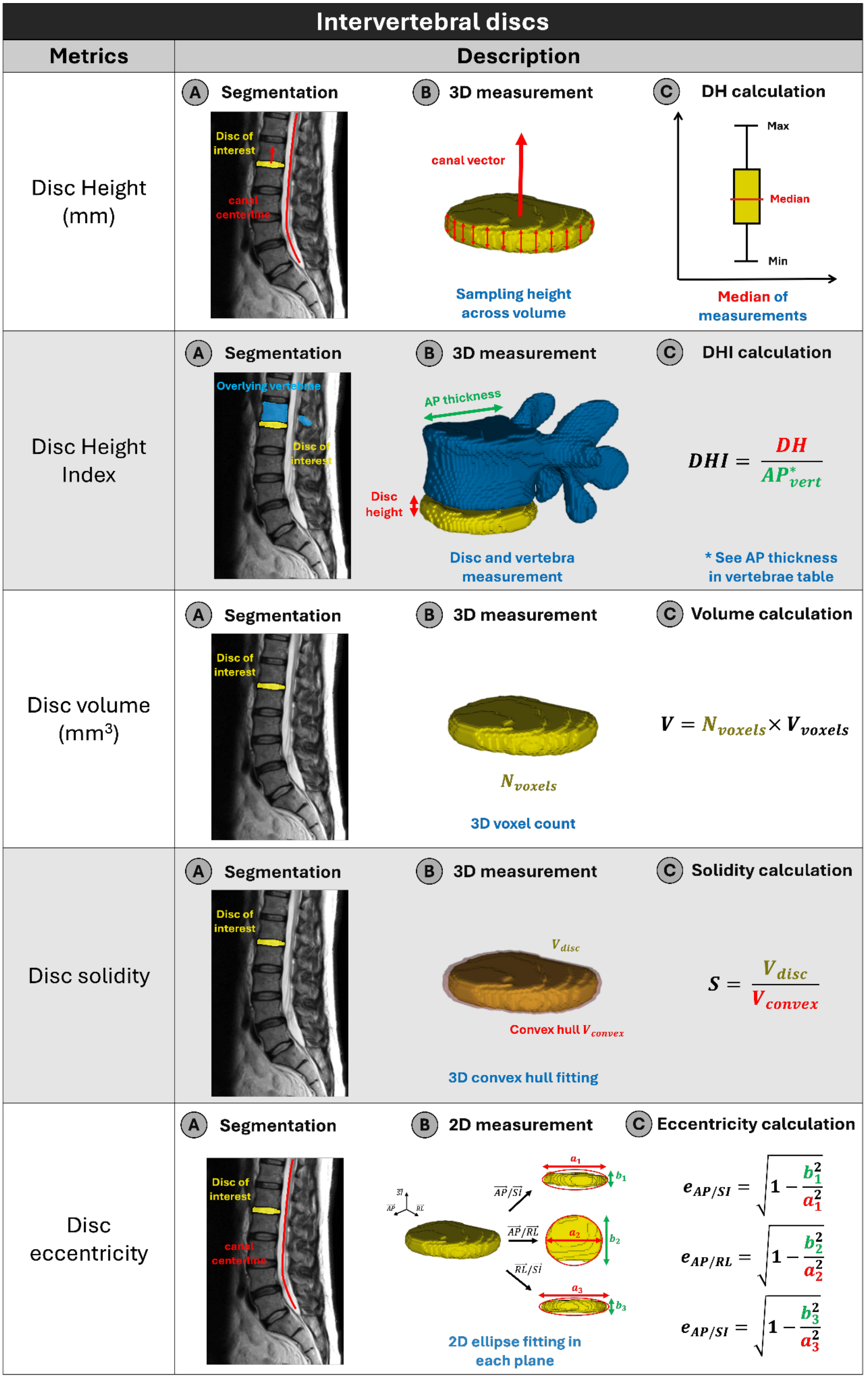

Intervertebral discs
Metrics
Description
Disc Height (mm)
A Segmentation
Disc of interest
canal centerline
B 3D measurement
canal vector
Sampling height across volume
C DH calculation
Max
Median
Min
Median of measurements
Disc Height Index
A Segmentation
Overlying vertebrae
Disc of interest
B 3D measurement
AP thickness
Disc height
Disc and vertebra measurement
C DHI calculation
$DHI = \frac{DH}{AP^{*}_{vert}}$
* See AP thickness in vertebrae table
Disc volume (mm³)
A Segmentation
Disc of interest
B 3D measurement
$N_{voxels}$
3D voxel count
C Volume calculation
$V = N_{voxels} \times V_{voxels}$
Disc solidity
A Segmentation
Disc of interest
B 3D measurement
$V_{disc}$
Convex hull $V_{convex}$
3D convex hull fitting
C Solidity calculation
$S = \frac{V_{disc}}{V_{convex}}$
Disc eccentricity
A Segmentation
Disc of interest
canal centerline
B 2D measurement
AP/SI
AP/RL
RL/SI
$a_1$ $b_1$ $a_2$ $b_2$ $a_3$ $b_3$
2D ellipse fitting in each plane
C Eccentricity calculation
$e_{AP/SI} = \sqrt{1 - \frac{b_1^2}{a_1^2}}$
$e_{AP/RL} = \sqrt{1 - \frac{b_2^2}{a_2^2}}$
$e_{AP/SI} = \sqrt{1 - \frac{b_3^2}{a_3^2}}$

**Figure 5.** Morphological measurements of the IVD illustrating how the disc height (DH), disc height index (DHI), volume, eccentricity, and solidity are derived from the segmentation.

In addition to morphological measurements, signal-based features of the IVD were extracted from T2-weighted MRI and normalized using the CSF signal as a reference. Following the approach of Waldenberg et al. (2018), the intensity histogram within the disc segmentation is analyzed to characterize disc health by separating low- and high-intensity components. Instead of Gaussian mixture modeling, we define these components using the 10th and 90th percentiles of the intensity distribution, which provides greater robustness across heterogeneous MRI protocols and avoids unstable peak fitting.

## 2.4 Reports generation

When *SpineReport* is applied to a cohort of patients, all previously computed morphometric metrics are aggregated on a per-subject basis, enabling group-level comparisons. For each anatomical structure, subject-specific measurements are aligned across the cohort by level for vertebrae, IVD, and IVF. For the spinal canal, point-wise labels at the posterior tip of the discs and the centerline length are used for each subject to resample, scale, and interpolate morphometric measurements to a common reference space, allowing direct comparison across patients.

Based on these aligned measurements, subject-specific reports are automatically generated to provide a comprehensive overview of spinal morphology (see **Figure 7**).

For IVDs, when MRI contrasts sensitive to water content are available, DHI and the low- and high-intensity components of the signal intensity are used and compared against the group distribution to automatically estimate relative modified Pfirrmann degeneration grades from 1 to 8 (Griffith et al. 2007). As with CSF measurements, these grades are relative and should be interpreted only in comparison with scans acquired using similar sequences and scanning conditions to avoid misinterpretation.

## 2.5 Lumbar pathologies evaluation

### 2.5.1 Lumbar pathologies

To assess the clinical relevance of the extracted morphometrics, all metrics were evaluated for their ability to discriminate three lumbar pathologies: lumbar central canal stenosis, lateral recess stenosis and foraminal stenosis. Each pathology was graded on a 0–3 severity scale. **Figure 6** displays grading examples for the three pathologies.

**Lumbar central canal stenosis (LCCS)** was graded following the work of Guen et al. (2011):
- Grade 0: no stenosis, with preserved anterior CSF space;
- Grade 1: mild stenosis, with slight CSF space obliteration but clear separation of cauda equina roots;
- Grade 2: moderate stenosis, with partial CSF obliteration and aggregation of some nerve roots;
- Grade 3: severe stenosis, with complete CSF obliteration and marked dural sac compression, where nerve roots appear as a single bundle.

**Lateral recess stenosis** was graded based on the work of Bartynski and Lin (2003):
- Grade 0: normal recess with preserved CSF and unaffected nerve root;
- Grade 1: mild narrowing with partial CSF reduction and no nerve root deformation;
- Grade 2: moderate narrowing with nerve root compression (flattening, widening, or deviation) but incomplete obliteration;
- Grade 3: severe narrowing with complete CSF obliteration and marked nerve root compression and displacement.

**Foraminal stenosis** was graded according to Lee et al. (2010):
- Grade 0: normal, with visible nerve root surrounded by epidural fat;
- Grade 1: mild stenosis with partial perineural fat obliteration but no nerve root deformation;
- Grade 2: moderate stenosis with complete fat obliteration around the nerve root without morphological changes;
- Grade 3: severe stenosis with nerve root collapse or morphological alteration due to pronounced structural narrowing.

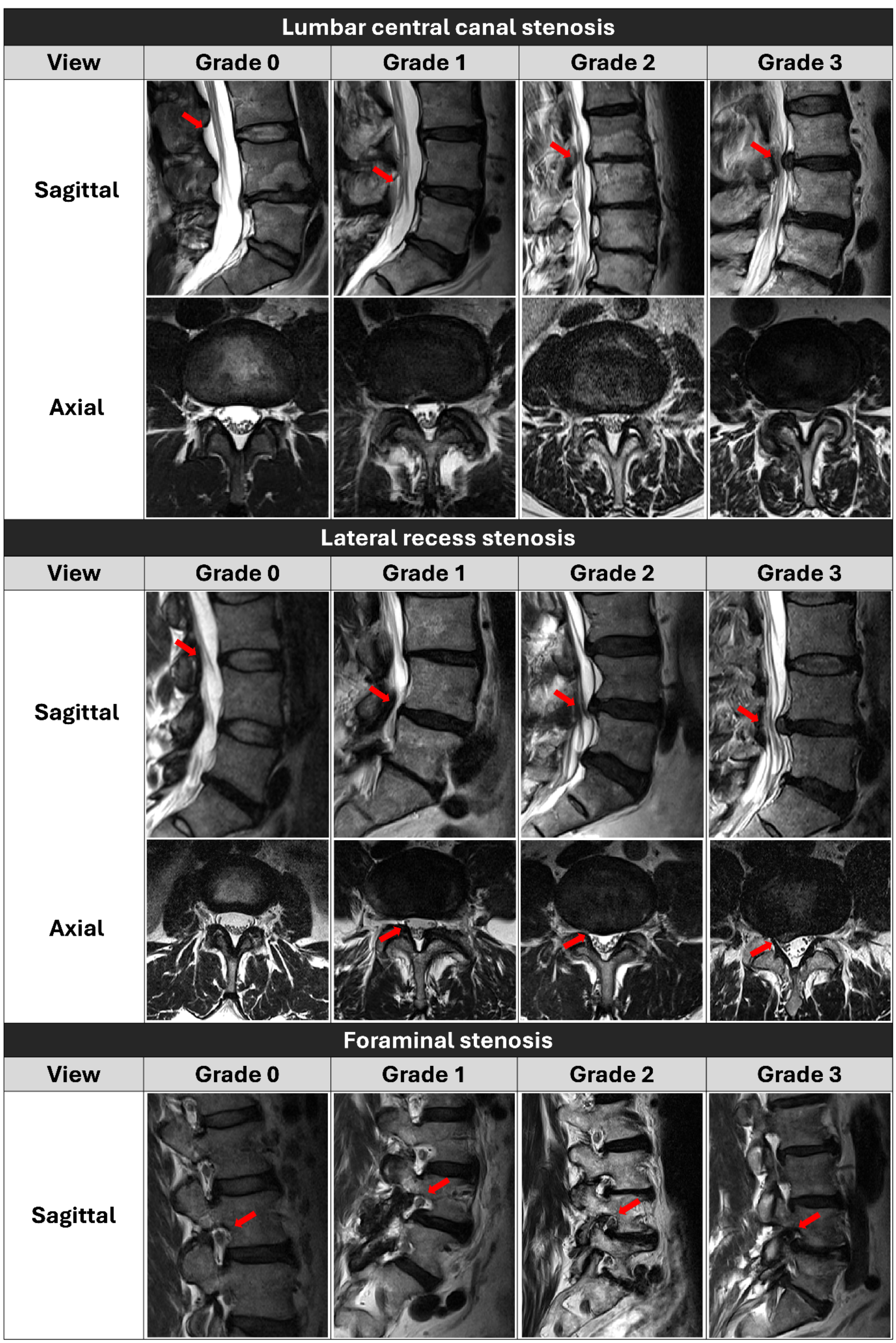

Lumbar central canal stenosis
View
Grade 0
Grade 1
Grade 2
Grade 3
Sagittal
Axial
Lateral recess stenosis
View
Grade 0
Grade 1
Grade 2
Grade 3
Sagittal
Axial
Foraminal stenosis
View
Grade 0
Grade 1
Grade 2
Grade 3
Sagittal

**Figure 6.** Examples of severity grading (0–3) for the three lumbar pathologies: central canal stenosis, lateral recess stenosis, and foraminal stenosis. When applicable, both sagittal and axial views are shown, except for foraminal stenosis, which is primarily assessed on sagittal views due to limited visibility on axial images. Red arrows indicate the graded level and side.

### 2.5.2 Dataset and demographics

A total of 272 patients (128 males, 144 females) were included, each presenting at least one lumbar pathology (lateral recess stenosis, foraminal stenosis, or central canal stenosis). The age ranged from 16 to 90 years (mean: 56.7 ± 14.7).

Sagittal T2-weighted, sagittal T1-weighted, and 1 to 5 axial T2-weighted scans were acquired using fast spin-echo MRI across all patients on Siemens (n=222), Philips (n=40), and GE (n=10) scanners, with field strengths of 0.34T (n=1), 1T (n=1), 1.5T (n=189), and 3T (n=81). The sagittal spatial resolution (LAS+) was 4.4 ± 0.46 mm (L), 0.58 ± 0.09 mm (A), and 0.58 ± 0.09 mm (S), while the axial resolution was 0.49 ± 0.11 mm (L), 0.49 ± 0.11 mm (A), and 4.2 ± 0.44 mm (S).

While sagittal scans covered the lower thoracic to sacral regions, axial scans had a limited field of view, each targeting individual disc levels. For each subject, one intervertebral level was graded by two board-certified musculoskeletal radiologists: AAM (5 years of experience, junior reader) and CG (>10 years, senior reader): L1–L2 (n=5), L2–L3 (n=18), L3–L4 (n=38), L4–L5 (n=124), and L5–S1 (n=93). Inter-rater agreement was substantial (Cohen's κ = 0.77), and the scores were averaged for subsequent analyses.

### 2.5.3 Spinal canal metrics post-processing

To align with disc-level grading by radiologists, slice-wise canal and CSF metrics from *SpineReport* were normalized by computing ratios between the average maximum values at the adjacent superior and inferior vertebrae and the minimum value within three millimeters around the disc level. This approach captures relative compression while accounting for inter-individual variability in canal size.

### 2.5.4 Measure stability

To assess the robustness and repeatability of morphometric extraction, measurements were computed from both sagittal T2-weighted and T1-weighted scans acquired during the same session. Agreement was evaluated using the intraclass correlation coefficient (ICC), Bland-Altman analysis, and Pearson correlation. This analysis was performed across aligned structures for spinal canal, IVD, vertebral body, and IVF morphometrics

### 2.5.5 Statistical analysis for pathology severity detection

Statistical analysis of associations with pathology severity included Spearman rank correlation to assess monotonic relationships without assuming linearity or normality, ordinal logistic regression to evaluate associations with increasing severity, and area under the curve (AUC) to measure discriminative performance between normal/mild and moderate/severe cases. P-values were adjusted using the Benjamini-Hochberg procedure to control the false discovery rate. All analyses were performed in Python 3.10 using pandas 2.3.3, SciPy 1.15.3, statsmodels 0.14.6, and NumPy 2.2.6.

### 2.5.6 Benchmarking against open-access approaches

Our approach was compared with another open-access tool, *SpineNet* (Windsor et al. 2024), which was developed to automatically assess degenerative changes at the disc level across several pathologies, including central canal and foraminal stenosis. *SpineNet* performs spinal canal stenosis grading using four severity levels and evaluates foraminal stenosis as a binary classification (presence vs. absence) more detailed descriptions of the severity grading criteria were not available. Adapting the tool to the data we used was done empirically including Reorientation and resampling to maximize the performances. Due to insufficient documentation, other open-access methods could not be included in this comparison (Hallinan et al. 2021; Batra et al. 2025).

# 3. Results

## 3.1 Example report

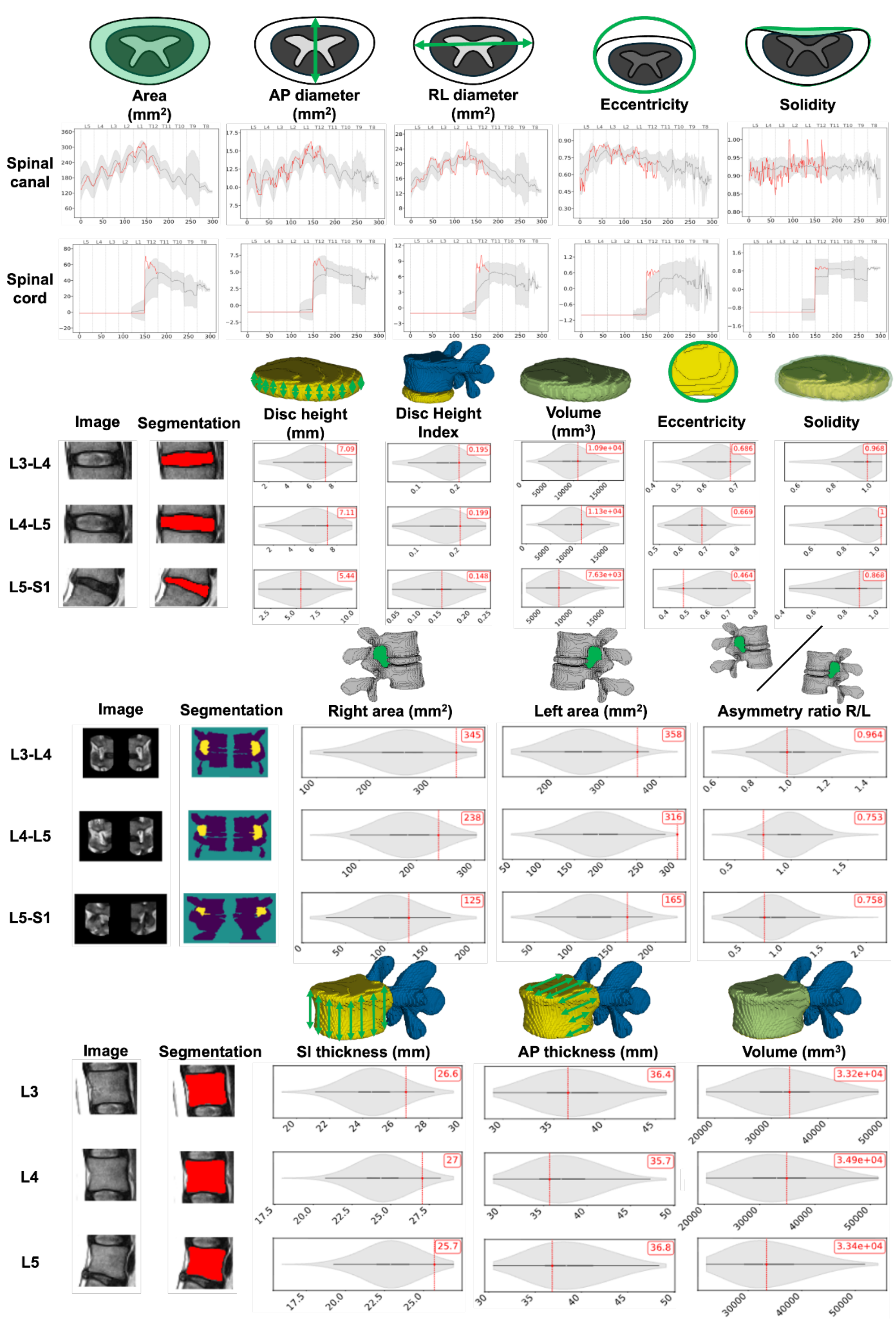

**Figure 7.** Example of a patient-specific report showing plots for the spinal canal, the spinal cord, vertebrae, IVF and IVD. The red lines represent the patient's values, while the gray distributions reflect cohort-level trends at the same anatomical level across the 272 subjects; deviations of the red line from the gray distribution highlight anomalies. The full report is available here:
https://github.com/ivadomed/SpineReport/blob/main/SpineReport/resources/imgs/example_report.pdf

## 3.2 Measurement stability

Morphometric measurements demonstrated high reproducibility between acquisitions, with strong agreement across most anatomical structures: For intervertebral discs, volumetric and thickness-related metrics showed excellent agreement, with intraclass correlation coefficients (ICC) of 0.98 for volume, 0.96 for median thickness, and 0.95 for DHI. Shape descriptors were also robust, including AP-RL eccentricity (ICC ≈ 0.87), RL-SI eccentricity (ICC ≈ 0.83) and solidity (ICC = 0.85), although eccentricity in the AP–SI plane showed moderately lower agreement (ICC = 0.67).

For vertebral bodies, volume and AP thickness exhibited excellent reproducibility (ICC = 0.96 for both), while median thickness showed slightly lower but acceptable agreement (ICC = 0.78).

For intervertebral foramina, left and right areas were highly reproducible (ICC = 0.94), whereas the asymmetry ratio showed moderate agreement (ICC = 0.58), likely reflecting exemplified variations caused by variations in the area measurements.

For the spinal canal, area and diameter measurements demonstrated excellent reproducibility, with ICC values of 0.98 (area), 0.98 (AP diameter), and 0.97 (RL diameter). In contrast, shape descriptors such as eccentricity and solidity were less stable (ICC = 0.30 and 0.48, respectively), indicating reduced robustness.

Bland–Altman plots for all morphological measurements of the spinal canal, IVD, vertebral bodies, and foramina are provided in the supplementary material (**Figure S1**).

## 3.3 Statistical evaluation

### 3.3.1 Sagittal versus axial analysis

Due to the limited field of view of the axial scans, morphometric extraction was not reliable. Key structures were often incomplete (e.g., foramina), and the small number of slices in the SI direction hindered robust centerline estimation. The field of view, centered on the disc, also resulted in incomplete vertebrae, preventing accurate vertebral body measurements. Additionally, the larger slice thickness in the SI direction reduced the reliability of disc metrics (e.g., disc height) due to increased partial volume effects, which are more pronounced given the smaller SI dimension of the disc compared to its AP and RL dimensions in sagittal views. Due to these limitations, severity scores were compared with continuous measurements extracted solely from the sagittal scans.

### 3.3.2 Lumbar central canal stenosis

The proposed morphological and T2-weighted signal-based metrics showed strong and significant associations with stenosis severity, along with high discriminative performance. The most predictive feature was the CSF signal at the graded level, with a strong negative Spearman correlation ($\rho = -0.61$, 95% CI [-0.68, -0.53], $p < 10^{-27}$) and excellent discrimination between normal/mild and moderate/severe cases (AUC = 0.95, 95% CI [0.95, 0.96], $p < 10^{-16}$). Ordinal logistic regression (OR = 0.14) indicated that a 23% decrease in CSF intensity was associated with an 86% increase in the likelihood of severe stenosis (95% CI [0.10, 0.20], $p < 10^{-24}$).

Similarly, the canal AP diameter ratio and canal cross-sectional area ratio were strongly correlated with severity ($\rho = -0.59$ and -0.56; 95% CI [-0.66, -0.51] and [-0.64, -0.48], respectively; $p < 10^{-22}$), with high discriminative performance (AUC = 0.85 and 0.80; 95% CI [0.84, 0.86] and [0.79, 0.80], respectively; $p < 10^{-7}$). Ordinal logistic regression (OR = 0.26) showed that a ~14% reduction in these normalized metrics relative to adjacent vertebral levels was associated with a 74% increase in the likelihood of severe stenosis (95% CI [0.20, 0.35], $p < 10^{-18}$).

The **Figure S2** A) and B) illustrates the progression of canal area and CSF signal in a case of grade 3 lumbar central canal stenosis at L2–L3 for the same patient.

### 3.3.3 Lateral recess stenosis

The proposed T2-weighted signal-based metrics showed significant but more moderate associations and discriminative performance. The most informative feature was the lateral CSF signal, which demonstrated a moderate negative correlation with severity ($\rho = -0.48$, 95% CI [-0.56, -0.38], $p = 1.86 \times 10^{-15}$) and good discriminative ability (AUC = 0.73, 95% CI [0.72, 0.74], $p = 4.42 \times 10^{-6}$). Ordinal logistic regression (OR = 0.41) indicated that a 23% decrease in lateral CSF intensity was associated with a 59% increase in the likelihood of severe stenosis (95% CI [0.32, 0.52], $p < 10^{-10}$).

The **Figure S2** C) and D) illustrate the progression of CSF right and left signal in a case of grade 3 left recess stenosis at L4–L5 for the same patient.

### 3.3.4 Foraminal stenosis

The proposed morphometrics did not show significant associations with severity scores. Although the region of interest was robustly extracted and integrated into subject-specific reports—facilitating automated and rapid level-wise evaluation—further assessment of stenosis severity was limited. This is likely due to the more complex image interpretation required, as simple signal intensity extraction was insufficient, particularly without reliable differentiation between nerve and fat signals.

The **Figure S3** illustrates the automatic extraction of intervertebral foramina across the four severity grades annotated by the senior reader.

### 3.3.5 Benchmarking against open-access approaches

In comparison, automated grading from *SpineNet* showed a weak and non-significant association with central canal stenosis. The Spearman correlation with severity grades was low ($\rho = 0.09$, 95% CI [−0.03, 0.21], $p = 0.13$), and discriminative performance was limited (AUC = 0.59, 95% CI [0.57, 0.60], $p = 0.06$). Ordinal logistic regression (OR = 1.18) indicated that a one-standard-deviation increase in the *SpineNet* score corresponded to an 18% increase in the likelihood of a higher stenosis grade, though this was not significant (95% CI [0.95, 1.48], $p = 0.13$).

For foraminal stenosis, *SpineNet* provides binary outputs indicating the presence or absence of pathology. Therefore, evaluation was limited to AUC analysis, using radiological grades as a continuous/ordinal predictor and SpineNet scores as the binary target. The method demonstrated limited discriminative performance (AUC = 0.65, 95% CI [0.64, 0.66], $p = 1.73 \times 10^{-5}$).

No grading was available for lateral recess stenosis, as *SpineNet* was not designed to assess this pathology.

# 4. Discussion

In this study, we introduced *SpineReport*, an automated deep-learning framework for the 3D quantification of spinal anatomy. By leveraging a robust segmentation pipeline, the method extracts both morphological and signal-based metrics to characterize spinal degeneration from 3D MRI scans.

Results demonstrated high measurement stability across T1-weighted and T2-weighted sequences for primary morphological measurements such as disc height, volume, and spinal canal cross-sectional area. In contrast, shape-related metrics, including eccentricity and solidity, showed lower stability—particularly for the spinal canal—likely reflecting greater sensitivity to minor variations. For example, despite high ICCs for AP and RL diameters, small measurement differences substantially impacted derived metrics such as canal eccentricity.

To evaluate clinical utility, we compared extracted metrics against semiquantitative radiological severity grades of lumbar spine conditions. Morphological and signal-based features showed significant associations with pathology, effectively discriminating between normal and severe central and recess stenosis (**Figure S2**). Notably, signal-based features of the CSF revealed pronounced hypo-intensities in severe cases, likely reflecting flow artifacts or intensity variations caused by high-grade stenosis. On the other hand, while signal intensity alone was insufficient to statistically characterize foraminal stenosis due to the complex differentiation between nerve and fat, our robust nerve localization enhances the clinical report (**Figure 7**), by making the foramina extraction automatic.

Several factors may have influenced the relationship between metric accuracy and severity grading, such as patient positioning, rater expertise, MRI artifacts or field inhomogeneities. As *SpineReport* relies on underlying anatomical segmentations, errors in the *TotalSpineSeg* output may propagate to the derived metrics. Moreover, while isotropic resampling (1 $mm^3$) standardizes the extraction process, it cannot recover anatomical details lost in low-resolution acquisitions, particularly in anisotropic data such as sagittal scans with lower RL resolution. For example, in foraminal stenosis, limited resolution affected the extraction of the region of interest, where the nerve root and surrounding fat were not always clearly distinguishable, complicating the analysis and interpolation could not compensate for these missing details. Additionally, while discrete grading is a valuable clinical tool for assessing disease severity and guiding treatment decisions, it provides only a coarse representation of structural changes. Variability within each grade cannot be fully captured by a four-level scale, whereas continuous morphometric measures can better reflect these subtle differences. This likely explains why Spearman correlations may be moderate while discriminative performance (e.g., AUC) remains high, as observed for CSF signal in central canal stenosis. Finally, signal-based metrics are more sensitive to image noise and acquisition variability than geometric features, making them more dependent on clinical imaging conditions and potentially less consistent across datasets.

Compared to *SpineNet*, *SpineReport* offers several key advantages. Although not specifically trained for individual pathologies, its morphometric and signal-based features enable continuous tracking of degenerative changes across spinal conditions. In contrast, *SpineNet* is constrained by its training design, and extending it to additional pathologies requires retraining, limiting its ability to support tasks such as lateral recess stenosis grading. *SpineReport* also showed stronger correlations with expert severity grading for both central canal and lateral recess stenosis. *SpineNet* showed weak discriminative performance for foraminal stenosis, and its binary output prevents a more granular evaluation of disease severity. In contrast, although *SpineReport* did not show direct correlations with extracted morphometrics, its automated extraction and reporting of the foraminal region of interest facilitate rapid visual assessment and improve efficiency for expert readers. Finally, comparisons with other open-access tools were limited by their availability and the quality of documentation, restricting comprehensive benchmarking across all three pathologies.

Beyond addressing the high workload of lumbar spine reporting, *SpineReport* provides a standardized platform for objective grading and longitudinal tracking of degeneration. While not a primary diagnostic tool, it offers clinicians critical quantitative markers and visualizations that identify anomalies, potentially accelerating intervention and improving diagnostic confidence. Future work will focus on improving the robustness of shape-based metrics, such as canal eccentricity and solidity, across imaging sequences, as well as developing additional quantitative biomarkers for foraminal stenosis, such as nerve root compression ratios as done by Gavotto et al. (2025). Extension to other lumbar pathologies not covered in this study, including disc bulging and scoliosis, is also planned.

# 5. Conclusion

This study introduces *SpineReport*, a robust framework for extracting morphological and signal-based metrics of spinal anatomy, including the spinal canal, the spinal cord, vertebrae, IVF and IVD. The tool provides quantitative 3D morphometrics and detailed subject-specific reports, enabling assessment against normative references and longitudinal analysis. Although not specifically designed for pathology diagnosis, *SpineReport* demonstrated moderate to strong and significant associations between its morphometrics and stenosis severity for both central canal and lateral recess stenosis, as well as robust extraction of regions of interest for foraminal stenosis. The tool is freely available (https://ivadomed.github.io/SpineReport/), facilitating broad accessibility and integration into existing neuroimaging pipelines.

# 6. Acknowledgements

This work was supported by the UNIQUE Excellence Doctoral scholarship, the Doctoral scholarship of the Fonds de Recherche du Québec, the Canada Research Chair in Quantitative Magnetic Resonance Imaging [CRC-2020-00179], the Canadian Institute of Health Research [PJT-190258], the Canada Foundation for Innovation [32454, 34824], the Fonds de Recherche du Québec - Santé [322736, 324636], the Natural Sciences and Engineering Research Council of Canada [RGPIN-2019-07244], the Canada First Research Excellence Fund (IVADO and TransMedTech), Mila - Tech Transfer Funding Program, the German Society of Musculoskeletal Radiology (Deutsche Gesellschaft für muskuloskelettale Radiologie; DGMSR), the Congressionally Directed Medical Research Programs (CDMRP HT9425-24-1-0868). During the preparation of this work the authors used ChatGPT and Gemini in order to assist with refining professional English. After using this tool, the authors reviewed and edited the content as needed and take full responsibility for the content of the published article.

# 7. Declaration of interest

The Radiology Department of Balgrist University Hospital has an academic research collaboration with Siemens Healthineers and Bayer. Reto Sutter has a joint patent submission with Siemens Healthineers, and he is a course director of IDKD.

# 8. References

Bartynski, Walter S., and Luke Lin. 2003. “Lumbar Root Compression in the Lateral Recess: MR Imaging, Conventional Myelography, and CT Myelography Comparison with Surgical Confirmation.” *AJNR: American Journal of Neuroradiology* 24 (3): 348–60.

Bartynski, Walter S., and Kalliopi A. Petropoulou. 2007. “The MR Imaging Features and Clinical Correlates in Low Back Pain–Related Syndromes.” *Magnetic Resonance Imaging Clinics of North America*, The Lumbar Spine, vol. 15 (2): 137–54. https://doi.org/10.1016/j.mric.2007.01.010.

Batra, Arnesh, Arush Gumber, and Anushk Kumar. 2025. *M-SCAN: A Multistage Framework for Lumbar Spinal Canal Stenosis Grading Using Multi-View Cross Attention*. https://doi.org/10.48550/arXiv.2503.01634.

Chen, Jiasong, Linchen Qian, Linhai Ma, Timur Urakov, Weiyong Gu, and Liang Liang. 2024. “SymTC: A Symbiotic Transformer-CNN Net for Instance Segmentation of Lumbar Spine MRI.” *Computers in Biology and Medicine* 179 (September): 108795. https://doi.org/10.1016/j.compbiomed.2024.108795.

Chen, Xiaolong, Harvinder S. Sandhu, Jose Vargas Castillo, and Ashish D. Diwan. 2021. “The Association between Pain Scores and Disc Height Change Following Discectomy Surgery in Lumbar Disc Herniation Patients: A Systematic Review and Meta-Analysis.” *European Spine Journal* 30 (11): 3265–77. https://doi.org/10.1007/s00586-021-06891-4.

De Leener, Benjamin, Simon Lévy, Sara M. Dupont, et al. 2017. “SCT: Spinal Cord Toolbox, an Open-

Source Software for Processing Spinal Cord MRI Data." *NeuroImage* 145 (January): 24–43. https://doi.org/10.1016/j.neuroimage.2016.10.009.

Gavotto, Amandine, Denys Fontaine, Roxane Fabre, Stephane Litrico, and Antoine Gennari. 2025. "MRI Evaluation of Lumbar Foraminal Stenosis: Correlation between a New Quantitative Evaluation and the Qualitative Lee's Classification." *European Spine Journal: Official Publication of the European Spine Society, the European Spinal Deformity Society, and the European Section of the Cervical Spine Research Society* 34 (7): 2908–13. https://doi.org/10.1007/s00586-025-08779-z.

Ghobrial, George, and Christian Roth. 2025. "Deep Learning-Based Automated Segmentation and Quantification of the Dural Sac Cross-Sectional Area in Lumbar Spine MRI." *Frontiers in Radiology* 5 (March): 1503625. https://doi.org/10.3389/fradi.2025.1503625.

Gilchrist, Russel V., Curtis W. Slipman, and Sarjoo M. Bhagia. 2002. "Anatomy of the Intervertebral Foramen." *Pain Physician* 5 (4): 372–78.

Graaf, Jasper W. van der, Miranda L. van Hooff, Constantinus F. M. Buckens, et al. 2024. "Lumbar Spine Segmentation in MR Images: A Dataset and a Public Benchmark." *Scientific Data* 11 (1): 264. https://doi.org/10.1038/s41597-024-03090-w.

Griffith, James F., Yi-Xiang J. Wang, Gregory E. Antonio, et al. 2007. "Modified Pfirrmann Grading System for Lumbar Intervertebral Disc Degeneration." *Spine* 32 (24): E708. https://doi.org/10.1097/BRS.0b013e31815a59a0.

Guen, Young Lee, Woo Lee Joon, Seok Choi Hee, Oh Kyoung-Jin, and Sik Kang Heung. 2011. "A New Grading System of Lumbar Central Canal Stenosis on MRI: An Easy and Reliable Method." *Skeletal Radiology* 40 (8): 1033–39. https://doi.org/10.1007/s00256-011-1102-x.

Hallinan, James Thomas Patrick Decourcy, Lei Zhu, Kaiyuan Yang, et al. 2021. "Deep Learning Model for Automated Detection and Classification of Central Canal, Lateral Recess, and Neural Foraminal Stenosis at Lumbar Spine MRI." *Radiology* 300 (1): 130–38. https://doi.org/10.1148/radiol.2021204289.

Hartvigsen, Jan, Mark J. Hancock, Alice Kongsted, et al. 2018. "What Low Back Pain Is and Why We Need to Pay Attention." *Lancet (London, England)* 391 (10137): 2356–67. https://doi.org/10.1016/S0140-6736(18)30480-X.

Herzog, Richard, Daniel R. Elgort, Adam E. Flanders, and Peter J. Moley. 2017. "Variability in Diagnostic Error Rates of 10 MRI Centers Performing Lumbar Spine MRI Examinations on the Same Patient within a 3-Week Period." *The Spine Journal* 17 (4): 554–61. https://doi.org/10.1016/j.spinee.2016.11.009.

Lee, Seunghun, Joon Woo Lee, Jin Sup Yeom, et al. 2010. "A Practical MRI Grading System for Lumbar Foraminal Stenosis." *American Journal of Roentgenology* 194 (4): 1095–98. https://doi.org/10.2214/AJR.09.2772.

LewandrowskI, Kai-Uwe, Narendran Muraleedharan, Steven Allen Eddy, et al. 2020. "Feasibility of Deep Learning Algorithms for Reporting in Routine Spine Magnetic Resonance Imaging." Special Issue. *International Journal of Spine Surgery* 14 (s3): S86–97. https://doi.org/10.14444/7131.

Maher, Chris, Martin Underwood, and Rachelle Buchbinder. 2017. "Non-Specific Low Back Pain." *The*

*Lancet* 389 (10070): 736–47. https://doi.org/10.1016/S0140-6736(16)30970-9.

Möller, Hendrik, Robert Graf, Joachim Schmitt, et al. 2025. "SPINEPS—Automatic Whole Spine Segmentation of T2-Weighted MR Images Using a Two-Phase Approach to Multi-Class Semantic and Instance Segmentation." *European Radiology* 35 (3): 1178–89. https://doi.org/10.1007/s00330-024-11155-y.

Pfirrmann, Christian W. A., Alexander Metzdorf, Marco Zanetti, Juerg Hodler, and Norbert Boos. 2001. "Magnetic Resonance Classification of Lumbar Intervertebral Disc Degeneration." *Spine* 26 (17): 1873.

Purushottam, Kumar, Singh Suyash, Balabantaray Bunil Kumar, and Nayak Rajashree. 2026. "Automated Quantitative Analysis of the Lumbar Spine: A Comprehensive Approach." *Journal of Imaging Informatics in Medicine*, ahead of print. https://doi.org/10.1007/s10278-025-01517-3.

Qian, Linchen, Jiasong Chen, Linhai Ma, Timur Urakov, Weiyong Gu, and Liang Liang. 2025. "Attention-Based Shape-Deformation Networks for Artifact-Free Geometry Reconstruction of Lumbar Spine From MR Images." *IEEE Transactions on Medical Imaging* 44 (12): 5258–77. https://doi.org/10.1109/TMI.2025.3588831.

Salem, Saied, Afnan Habib, Mukhlis Raza, Ahmet Arif Aydin, Yeong Hyeon Gu, and Mugahed A. Al-Antari. 2025. "SpineAutoCAD: Multimodal CAD System for Lumbar Spine MRI Analysis and Structured Report Generation." September 6, 1–6. https://doi.org/10.1109/idap68205.2025.11222171.

Samartzis, Dino, Jaro Karppinen, Jason Pui Yin Cheung, and Jeffrey Lotz. 2013. "Disk Degeneration and Low Back Pain: Are They Fat-Related Conditions?" *Global Spine Journal* 3 (3): 133–44. https://doi.org/10.1055/s-0033-1350054.

Subramanian, Bargava, Naveen Kumarasami, Praveen Shastry, et al. 2025. "AI-Driven MRI Spine Pathology Detection: A Comprehensive Deep Learning Approach for Automated Diagnosis in Diverse Clinical Settings." arXiv:2503.20316. Preprint, arXiv, March 28. https://doi.org/10.48550/arXiv.2503.20316.

Tunset, Andreas, Per Kjaer, Shadi Samir Chreiteh, and Tue Secher Jensen. 2013. "A Method for Quantitative Measurement of Lumbar Intervertebral Disc Structures: An Intra- and Inter-Rater Agreement and Reliability Study." *Chiropractic & Manual Therapies* 21 (1): 26. https://doi.org/10.1186/2045-709X-21-26.

Udomluck, Phatcharapon, Watcharaporn Cholamjiak, Jakkaphong Inpun, and Waragunt Waratamrongpatai. 2026. "Multi-Task Deep Learning Model for Automated Detection and Severity Grading of Lumbar Spinal Stenosis on MRI: Multi-Center External Validation." *Diseases* 14 (1): 32. https://doi.org/10.3390/diseases14010032.

Waldenberg, Christian, Hanna Hebelka, Helena Brisby, and Kerstin Magdalena Lagerstrand. 2018. "MRI Histogram Analysis Enables Objective and Continuous Classification of Intervertebral Disc Degeneration." *European Spine Journal* 27 (5): 1042–48. https://doi.org/10.1007/s00586-017-5264-7.

Wang, Aobo, Tianyi Wang, Xingyu Liu, et al. 2025. "Automated Diagnosis and Grading of Lumbar Intervertebral Disc Degeneration Based on a Modified YOLO Framework." *Frontiers in*

*Bioengineering and Biotechnology* 13 (January): 1526478. https://doi.org/10.3389/fbioe.2025.1526478.

Warszawer, Yehuda, Nathan Molinier, Jan Valosek, et al. 2025. "TotalSpineSeg: Robust Spine Segmentation with Landmark-Based Labeling in MRI." Preprint, ResearchGate. https://doi.org/10.13140/RG.2.2.31318.56649.

Windsor, Rhydian, Amir Jamaludin, Timor Kadir, and Andrew Zisserman. 2022. "Context-Aware Transformers For Spinal Cancer Detection and Radiological Grading." arXiv:2206.13173. Preprint, arXiv, June 27. https://doi.org/10.48550/arXiv.2206.13173.

Windsor, Rhydian, Amir Jamaludin, Timor Kadir, and Andrew Zisserman. 2024. "Automated Detection, Labelling and Radiological Grading of Clinical Spinal MRIs." *Scientific Reports* 14 (1): 14993. https://doi.org/10.1038/s41598-024-64580-w.

Zheng, Hua-Dong, Yue-Li Sun, De-Wei Kong, et al. 2022. "Deep Learning-Based High-Accuracy Quantitation for Lumbar Intervertebral Disc Degeneration from MRI." *Nature Communications* 13 (1): 841. https://doi.org/10.1038/s41467-022-28387-5.

# 8. Supplementary

## 8.1 Measurement stability

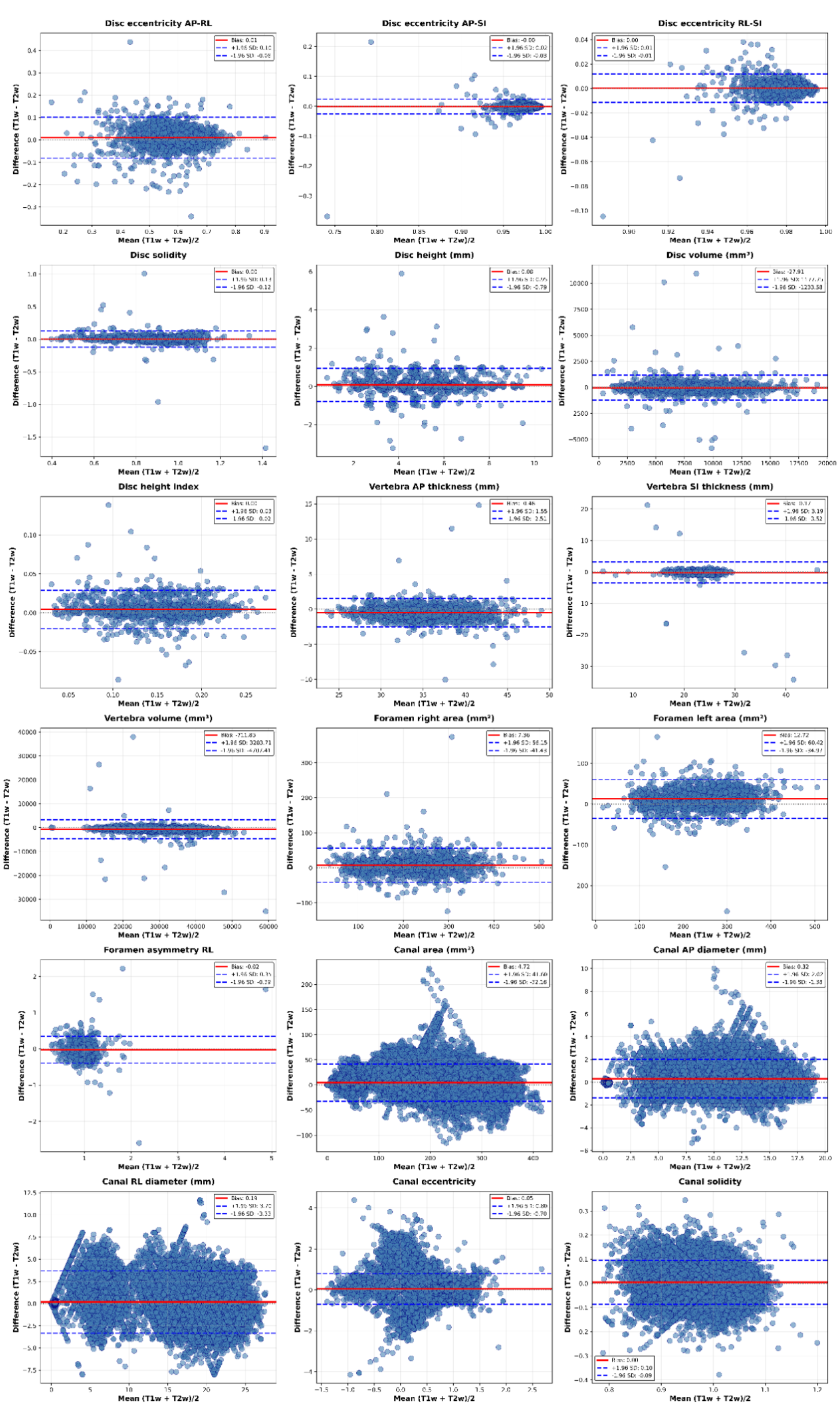

**Figure S1.** Bland–Altman plots (95% limits of agreement) comparing T1-weighted and T2-weighted measurements for spinal canal, IVD, IVF, and vertebrae extracted with SpineReport across 272 subjects.

## 8.2 Statistical evaluation

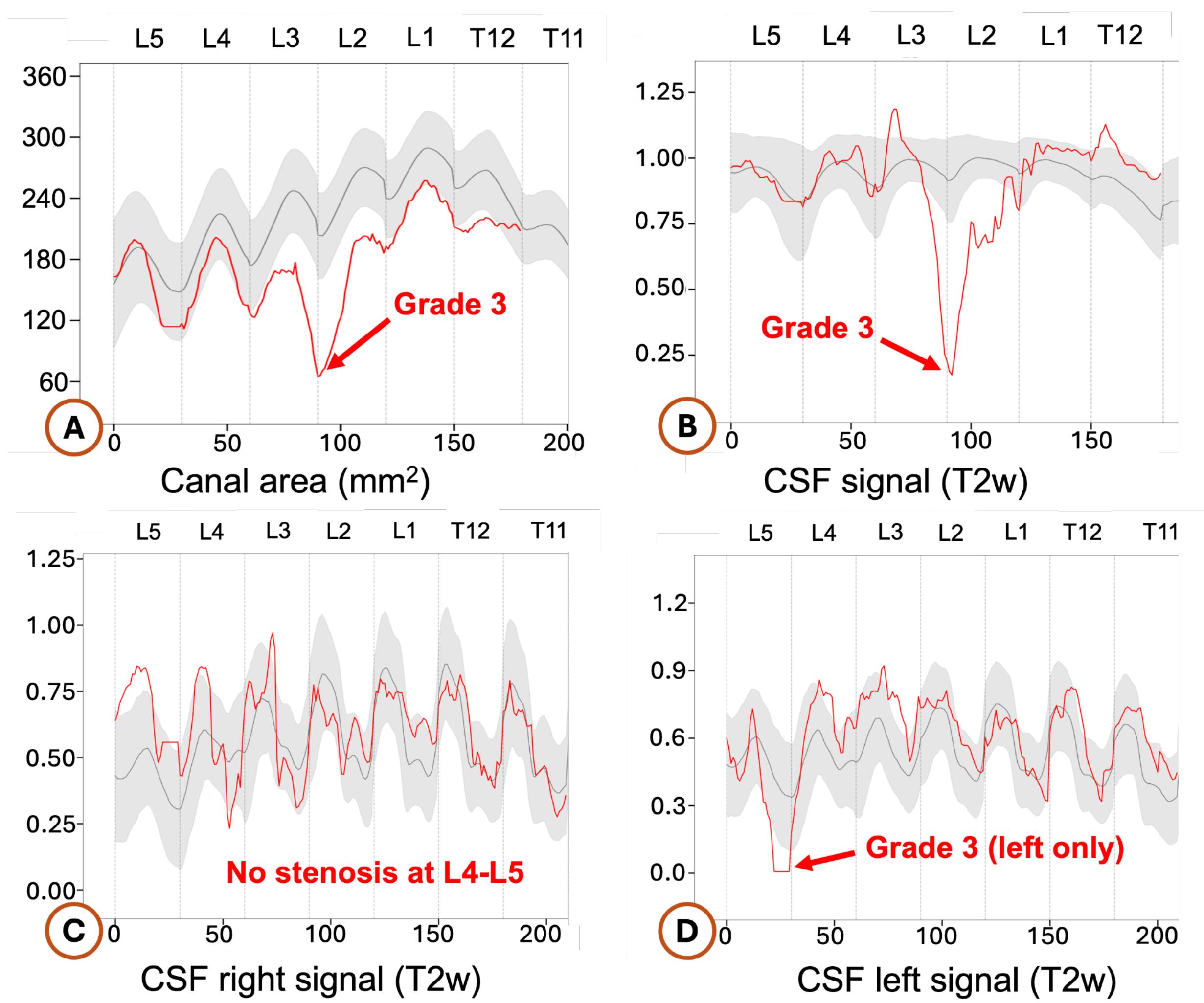

**Figure S2.** Morphometric measurements for two patients. (A) and (B) show canal area and T2-weighted CSF signal across superior–inferior slices for a patient with grade 3 lumbar central canal stenosis at L2–L3. (C) and (D) show right and left CSF signals across superior–inferior slices for a patient with grade 3 left lateral recess stenosis at L4–L5. The red lines represent the patient's values, while the gray distributions reflect cohort-level trends at the same anatomical level across the 272 subjects; deviations of the red line from the gray distribution highlight anomalies.

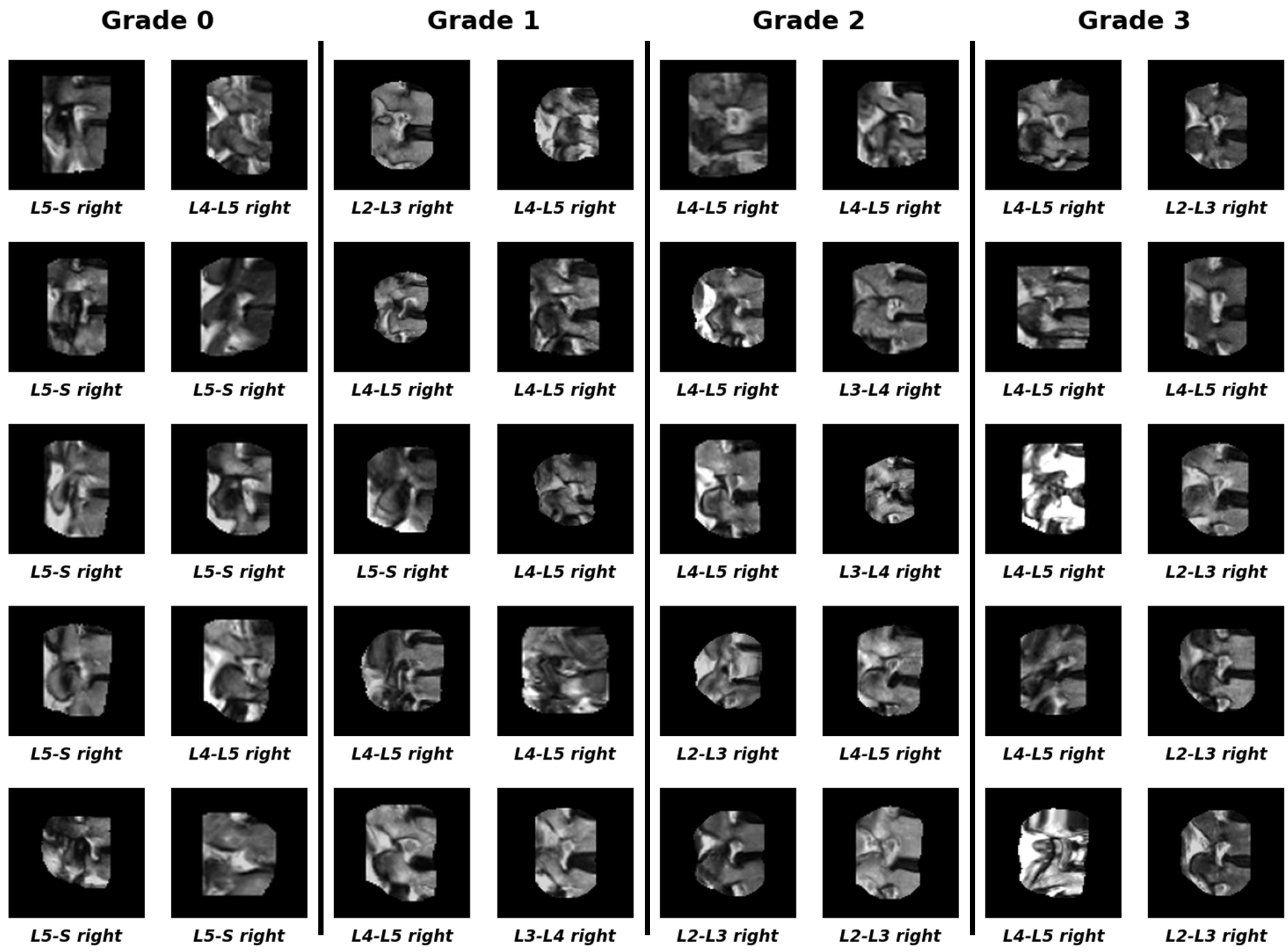

**Figure S3.** Automatic extraction of intervertebral foramina in ten patients at the specified level and side, illustrating the four foraminal stenosis severity grades annotated by the senior reader.